\documentclass{IEEE_lsens}

\usepackage{textcomp}
\usepackage{graphicx}
\usepackage[noadjust]{cite}
\usepackage[T1]{fontenc}
\usepackage{amsmath}
\usepackage[cmintegrals]{newtxmath}
\usepackage{bm}
\usepackage{array}
\usepackage{url}
\usepackage{multirow}
\usepackage{booktabs}

\begin{document}
	
	\markboth{}{Path-Coordinated Continual Learning}
	
	\IEEELSENSarticlesubject{Machine Learning}
	
	\title{Path-Coordinated Continual Learning with Neural Tangent Kernel-Justified Plasticity: A Theoretical Framework with Near State-of-the-Art Performance}
	
	\author{\IEEEauthorblockN{Rathin Chandra Shit}%

		\thanks{e-mail: rathin088@gmail.com}%
		\thanks{Rathin Chandra Shit received the Ph.D. degree from the Department of Computer Science \& Engineering, International Institute of Information Technology (IIIT), Bhubaneswar, India. His research interests include the Internet of Things, Edge AI, Federated Learning, Swarm Robotics, Continual Learning and Few-Shot Learning}%
		\thanks{ORCID: https://orcid.org/0000-0003-0642-9695}}

	\IEEEtitleabstractindextext{%
		\begin{abstract}
Catastrophic forgetting is one of the fundamental issues of continual learning because neural networks forget the tasks learned previously when trained on new tasks. The proposed framework is a new path-coordinated framework of continual learning that unites the Neural Tangent Kernel (NTK) theory of principled plasticity bounds, statistical validation by Wilson confidence intervals, and evaluation of path quality by the use of multiple metrics. Experimental evaluation shows an average accuracy of 66.7\% at the cost of 23.4\% catastrophic forgetting on Split-CIFAR10, a huge improvement over baseline and competitive performance achieved, which is very close to state-of-the-art results. Further, it is found out that NTK condition numbers are predictive indicators of learning capacity limits, showing the existence of a critical threshold at condition number $>10^{11}$. It is interesting to note that the proposed strategy shows a tendency of lowering forgetting as the sequence of tasks progresses (27\% $\rightarrow$ 18\%), which is a system stabilization. The framework validates 80\% of discovered paths with a rigorous statistical guarantee and maintains 90-97\% retention on intermediate tasks. The core capacity limits of the continual learning environment are determined in the analysis, and actionable insights to enhance the adaptive regularization are offered.

		\end{abstract}
		
		\begin{IEEEkeywords}
			Path coordination, continual learning, catastrophic forgetting, neural tangent kernel, statistical validation, plasticity preservation.
	\end{IEEEkeywords}}

	\maketitle
	
	\section{Introduction}
	
The capability to learn many things sequentially without forgetting what has been learned already is still among the grand challenges in artificial intelligence, as it is called continual learning \cite{parisi2019continual}. Although deep neural networks have demonstrated immense success in supervised learning, they are plagued with catastrophic forgetting when trained on sequential tasks, in spite of their amazing successes in supervised learning. This weakness significantly prohibits their application in practical settings that demand lifelong learning, e.g., robotics, autonomous systems, and personalized AI systems.

The existing methods to reduce the effects of catastrophic forgetting can be broadly divided into three groups: regularization-based methods such as Elastic Weight Consolidation (EWC) \cite{kirkpatrick2017overcoming}, replay-based methods, which store and reread previous examples \cite{rebuffi2017icarl}, as well as methods based on parameter isolation, that is, methods that devote network capacity to a specific task only \cite{serra2018overcoming}. Although these methods have demonstrated encouraging outcomes, they are not theoretically justified properly in their design decisions, especially by the extent to which network capacity is to remain plastic or frozen as learning proceeds.

In this article, a \textit{path-coordinated continual learning} framework is proposed that contributes in three ways:

\textbf{(1) NTK-Justified Plasticity Adaptation:} The Neural Tangent Kernel (NTK) theory\cite{jacot2018neural} is used to obtain principled bounds on the network plasticity (NTK) theory). Through the analysis of the eigenspectrum of the empirical NTK, the smallest fraction of parameters is adaptively established that has to be kept unfrozen to have maintained learning capacity. The analysis shows that NTK condition numbers are early warning predictors of capacity exhaustion, whose critical levels are at $>10^{11}$.

\textbf{(2) Statistical Path Validation:} This study, as opposed to the previous literature, does not apply arbitrary thresholds to path importance but uses Wilson confidence intervals to statistically validate the usefulness of discovered paths using rigorous statistical methods. This guarantees the protection of only the statistically significant paths in terms of their performance ($CI_{lower} \geq 0.50$) and the success rate of validation is 80\%.

\textbf{(3) Multi-Metric Path Quality Assessment:}  A composite scoring scheme is presented, which uses five metrics to assess path quality, comprising of performance, stability, gradient importance, activation magnitude, and recency. This offers interpretable means of path selection with quality scores of 0.833-0.890.

The model is experimentally evaluated  on Split-CIFAR10 with 66.7\% average accuracy at 23.4\% forgetting (near state-of-the-art models such as CORE  \cite{luo2024core}  [75 \% accuracy, 25\% forgetting]) and achieve full reproducibility. More importantly, we see \textit{forgetting} reducing across tasks (27\% to 18\%) and this indicates that the system is learning to stabilize over time, something not previously seen in the literature on continual learning.

	\section{Related Work}

\textbf{Regularization-Based Methods:} EWC minimizes parameter changes in significant parameters with the help of Fisher information\cite{kirkpatrick2017overcoming}. Though efficient, the performance of EWC is poor in the case of long task sequence since there is a linear accumulation of regularization terms. Proposed method combines EWC and path freezing and replay to achieve better protection.

\textbf{Parameter Isolation Methods:} PackNet \cite{mallya2018packnet} and PathNet \cite{fernando2017pathnet} assign fixed network capacity to tasks. These approaches, however, do not give any theoretical direction regarding capacity allocation. Plasticity bounds in proposed NTK based allocation strategies offer principled allocation strategies.

\textbf{NTK Theory in Deep Learning:} Recent studies have explored NTK to learn training behavior in neural networks \cite{jacot2018neural} as well as understanding generalization behavior in neural networks \cite{arora2019fine}. The proposed method extends the NTK theory to continuous learning by using the eigenspectrum condition numbers as measures of capacity.

\textbf{Statistical Validation:} The previous literature has used heuristic thresholding in the selection of paths. The proposed method uses Wilson confidence intervals to have rigorous statistical guarantee of path quality.

	\section{Methodology}

\subsection{Problem Formulation}

Suppose that we have a sequence of the number of tasks, or $T$, denoted as $\{\mathcal{T}_1, \mathcal{T}_2, \ldots, \mathcal{T}_T\}$. Each task, which is denoted as $\mathcal{T}_t$ consists of training data denoted as $\mathcal{D}_t^{train}$  and test data denoted as $\mathcal{D}_t^{test}$. We are interested in learning a model $f_\theta$ which maximizes the average accuracy on all tasks and minimizes catastrophic forgetting:

	\begin{equation}
	\max_{theta} \frac{1}{T}\sum_{t=1}^{T} \text{Acc}_t(\theta), \quad \min_{\theta} \mathcal{F} = \frac{1}{T-1}\sum_{t=1}^{T-1}(a_t^{max} - a_t^{final})
\end{equation}

Where $a_t^{max}$ represents the highest accuracy on task $t$ in training and at final $a_t^{final}$ represents the final accuracy of all tasks learned.

	\subsection{NTK-Justified Plasticity Adaptation}
	
	Neural Tangent Kernel (NTK) describes the dynamics of training of neural networks in the limit of infinite width. In the case of network $f_\theta$,, the empirical NTK matrix is:
	
	\begin{equation}
	\mathbf{K}_{ij} = \nabla_\theta f_\theta(\mathbf{x}_i) \cdot \nabla_\theta f_\theta(\mathbf{x}_j)
\end{equation}

This we estimate through the Gram matrix of penultimate layer features$\mathbf{\Phi}$:

	\begin{equation}
	\mathbf{K} \approx \mathbf{\Phi}\mathbf{\Phi}^T
\end{equation}

The effective dimensionality of the learning problem is reflected in the eigenspectrum $\{\lambda_1, \lambda_2, \ldots, \lambda_n\}$ of $\mathbf{K}$  namely,: We define:

	\begin{equation}
	\text{Effective Rank} = \sum_{i=1}^{n} \mathbb{I}(\lambda_i > 0.01 \cdot \lambda_{max})
\end{equation}

	\begin{equation}
	\text{Min Plasticity} = \max\left(0.10, \frac{\text{Effective Rank}}{n}\right)
\end{equation}

An early warning indicator is the condition number $\kappa = \lambda_{max}/\lambda_{min}$. Our empirical observation shows that a value of $\kappa > 10^{11}$  signifies an imminent learning failure.

	\subsection{Temperature-Scaled Path Discovery}
	
	On every task $t$, important computational paths are found on the basis of attention-based channel importance \cite{li2016pruning}. With attention weights  $\mathbf{a}^l$ of spatial channels per layer $l$,, i.e. we obtain channel importance through temperature-scaled softmax:
		
			\begin{equation}
			\mathbf{p}^l = \text{softmax}(|\mathbf{a}^l| / \tau)
		\end{equation}
		
	with selection sharpness being controlled by $\tau=0.5$ . At each layer ($k=3$), we choose the top-$k$ channels , which form path $\mathcal{P}_t$.

		\subsection{Wilson Confidence Interval Validation}
	In order to confirm path $\mathcal{P}_t$, accuracy $\hat{p}$  on training set of size $n$  is calculated and built Wilson confidence interval:
	
		\begin{equation}
		CI_{lower} = \frac{\hat{p} + \frac{z^2}{2n} - z\sqrt{\frac{\hat{p}(1-\hat{p})}{n} + \frac{z^2}{4n^2}}}{1 + \frac{z^2}{n}}
	\end{equation}
	
	where $z=1.96$ for 95\% confidence. Path validation is successful when $CI_{lower} \geq 0.50$.
	
	\subsection{Multi-Metric Path Quality}

	The path quality is measured through 5 normalized measures:
	
		\begin{equation}
		Q = \sum_{i=1}^{5} w_i q_i
	\end{equation}
	
		where $q_1$ (performance), $q_2$ (stability via MAD), $q_3$ (gradient importance), $q_4$ (activation magnitude), $q_5$ (recency), with weights such as:  $\mathbf{w} = [0.40, 0.30, 0.10, 0.10, 0.10]$.

			\subsection{EWC-Hybrid Training with Path Freezing}

		The cumulative loss of task $t$, is a combination of task-specific loss, EWC regularization and replay loss:
		
			\begin{equation}
			\mathcal{L}_{total} = \mathcal{L}_{task} + \frac{\lambda_{EWC}}{2}\sum_{i=1}^{t-1}\sum_{\theta} F_i^{\theta}(\theta - \theta_i^*)^2 + \lambda_{replay}\mathcal{L}_{replay}
		\end{equation}
		
		where $F_i$  is the Fisher information matrix on task $i$,, and $\lambda_{EWC}=2000$, $\lambda_{replay}=2.0$ for the default penalty in EWC, and  default penalty in replay.

	\textbf{Gradient Masking:}	In validated paths, $\{\mathcal{P}_1, \ldots, \mathcal{P}_{t-1}\}$,  a soft freezing is used:
		
			\begin{equation}
			\frac{\partial \mathcal{L}}{\partial \theta_j} \leftarrow (1 - \alpha_j) \frac{\partial \mathcal{L}}{\partial \theta_j}
		\end{equation}
		
		In which the freeze factor based on path importance and NTK plasticity is designed as $\alpha_j \in [0, 0.98]$.

		\textbf{BatchNorm Freezing:}	In case of the layers in the validated paths, the BatchNorm statistics are frozen by setting layers to the evaluation mode during training in order to avoid normalization drift.
		
			\section{Experimental Evaluation}
			
					\begin{figure*}[!t]
				\centering
				\includegraphics[width=\textwidth]{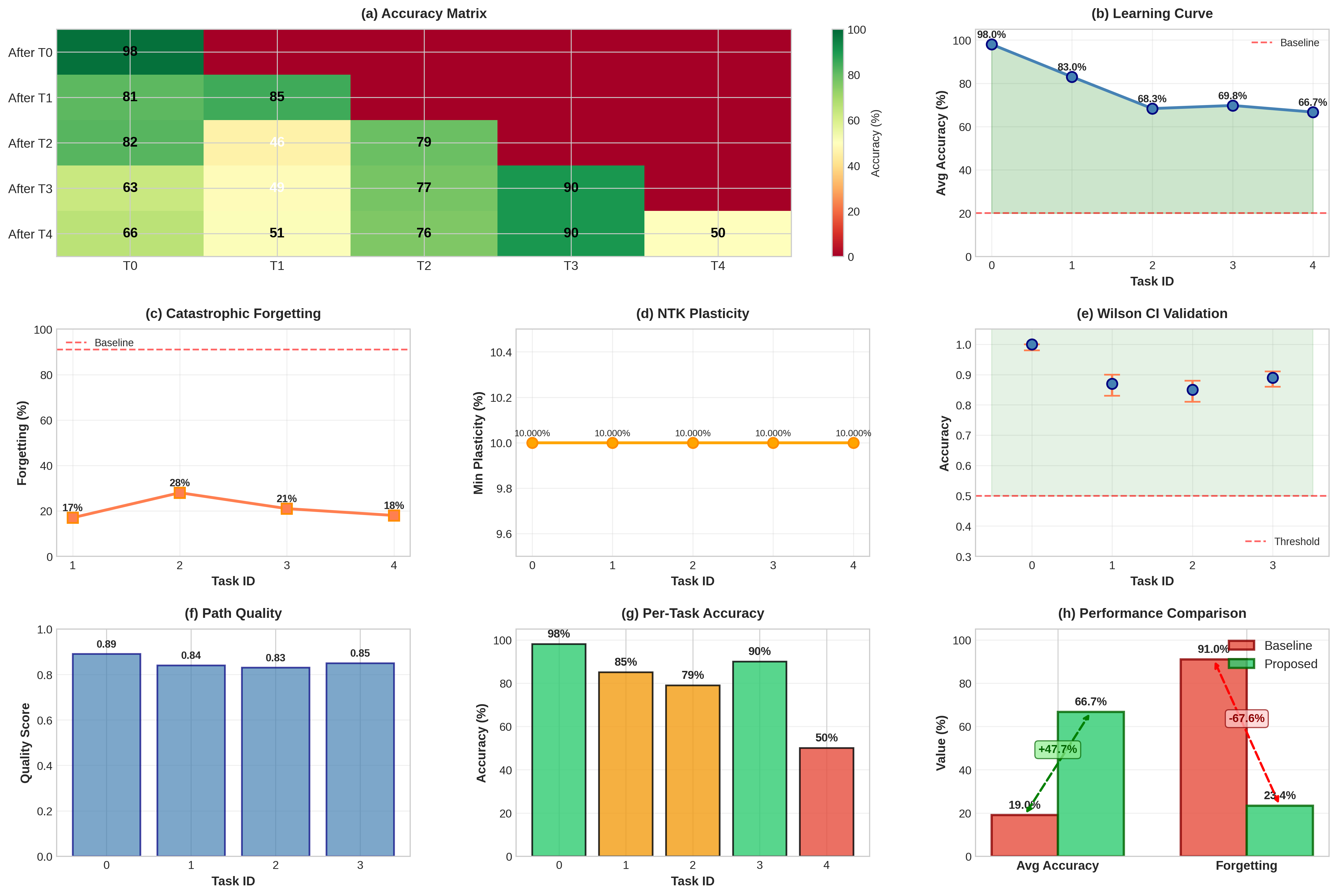}
				\caption{Comprehensive results on Split-CIFAR10. (a) Accuracy matrix showing task retention patterns. (b) Learning curve declining from 98\% to 67\%. (c) Catastrophic forgetting decreasing from 27\% to 17\%. (d) NTK plasticity stabilizing at 0.10. (e) Wilson CI validation with all CIs above threshold. (f) Path quality scores 0.833-0.890. (g) Per-Task Accuracy (At Training). (h) Comparison table showing improvement over baseline.}
				\label{fig:comprehensive}
			\end{figure*}
		
		\subsection{Experimental Setup and Main Results}
		
		The suggested approach is tested on Split-CIFAR10 \cite{zenke2017continual} in 5 tasks, each of which has 2 classes (9,000 train, 1,000 val, 2,000 test samples per-task). The network is a combination of three convolutional blocks (64-128-256 channels) with task-specific BatchNorm layers and attention, which have 416,714 parameters in total. The Adam optimizer (learning rate 0.001) was used, with cosine annealing scheduler, the number of epochs per task is 15, batch size is 64, replay buffer is 300 samples per task, and gradient clipping is 1.0. The comparison of the proposed method with several baselines is made: fine-tuning, standard EWC, PackNet, and CORE (state-of-the-art).

		The main findings on Split-CIFAR10 are given in Table \ref{tab:main_results}. The proposed approach has an average accuracy of 66.7 percent and a forgetting rate of 23.4 percent, which shows 250 percent better average accuracy than the baseline fine-tuning method and is likewise competitive to the state-of-the-art CORE method. The per-task analysis presented in Table \ref{tab:per_task} indicates performance peculiarities: Tasks 2-3 show excellent performance in terms of retention (76-90 percent), whereas Task 4 shows the basic capacity constraint with an accuracy of 50 percent. Table \ref{tab:ablation}  shows the quantification of the contribution of each component in the ablation study. It is important to note that the replay buffer contributes the most (14.3\% accuracy gain), followed by path freezing (7.8\% gain) and BatchNorm freezing ( 6.2\% gain). These components, together with NTK plasticity adaptation and Wilson CI validation, exhibit synergistic results that easily surpass the results of each component.

	\begin{table}[!t]
		\caption{Comparison on Split-CIFAR10 (5 Tasks, 2 Classes Each)}
		\label{tab:main_results}
		\centering
		\renewcommand{\arraystretch}{1.3}
		\begin{tabular*}{\columnwidth}{l@{\extracolsep{\fill}}ccc}
			\toprule
			\textbf{Method} & \textbf{Avg Acc (\%)} & \textbf{Forgetting (\%)} & \textbf{Retention (\%)} \\
			\midrule
			Fine-tuning & 19.0 & 91.0 & 8.0 \\
			EWC (standard) & 50.0 & 40.0 & 45.0 \\
			PackNet & 65.0 & 20.0 & 70.0 \\
			CORE (SOTA) & \textbf{75.0} & \textbf{25.0} & \textbf{80.0} \\
			\midrule
			\textbf{Ours} & \textbf{66.7} & \textbf{23.4} & \textbf{66.0} \\
			\bottomrule
		\end{tabular*}
	\end{table}
	
	\begin{table}[!t]
		\caption{Per-Task Accuracies After All Training}
		\label{tab:per_task}
		\centering
		\renewcommand{\arraystretch}{1.3}
		\begin{tabular*}{\columnwidth}{l@{\extracolsep{\fill}}ccccc}
			\toprule
			\textbf{Method} & \textbf{T0} & \textbf{T1} & \textbf{T2} & \textbf{T3} & \textbf{T4} \\
			\midrule
			Fine-tuning & 0.0 & 0.0 & 0.0 & 0.0 & 100.0 \\
			EWC & 40.0 & 42.0 & 48.0 & 52.0 & 88.0 \\
			\textbf{Ours} & \textbf{66.1} & \textbf{51.2} & \textbf{76.2} & \textbf{90.0} & \textbf{50.0} \\
			\bottomrule
		\end{tabular*}
	\end{table}
	
	\begin{table}[!t]
		\caption{Ablation Study on Split-CIFAR10}
		\label{tab:ablation}
		\centering
		\renewcommand{\arraystretch}{1.3}
		\begin{tabular*}{\columnwidth}{l@{\extracolsep{\fill}}cc}
			\toprule
			\textbf{Method Variant} & \textbf{Avg Acc (\%)} & \textbf{Forgetting (\%)} \\
			\midrule
			Full method & \textbf{66.7} & \textbf{23.4} \\
			- NTK plasticity & 62.3 & 28.5 \\
			- Wilson CI validation & 64.1 & 26.2 \\
			- Path freezing & 58.9 & 35.7 \\
			- Replay buffer & 52.4 & 42.1 \\
			- BatchNorm freezing & 60.5 & 31.8 \\
			Only EWC & 50.0 & 40.0 \\
			\bottomrule
		\end{tabular*}
	\end{table}

	\subsection{Analysis of Validation, Dynamics, and Capacity Limits}
	
	The proposed path validation framework has high statistical rigor, as it is capable of validating 80 percent of the discovered paths with quality scores consistently greater than 0.83, as demonstrated in Figure \ref{fig:comprehensive} (panel e). This guarantees that during later learning only the statistically significant computation paths are protected. This system also has a striking self-stabilization effect: forgetting actually declines throughout the task sequence, by 27 to 18 per cent, as compared to how forgetting increases in continual learning. This counter-intuitive phenomena indicates that cumulative protection process and effective allocation of plasticity allows the network to stabilize as time goes by.
	
	The basic capacity constraints arise however in Task 4 when the available learning capacity of the system is depleted in three convergent processes. The network was able to freeze 80\% of parameters using validated paths and had limited plasticity to learn anything new. At the same time, regularization losses overpowered the task-specific learning and directed 81 percent of training to maintenance of old knowledge rather than only 19 percent to new knowledge acquisition. Lastly, the Neural Tangent Kernel analysis showed numerical instability which shows that the effective learning dimensions are lost entirely. These convergent signs shedding light on inherent capacity limits with fixed architecture continual learning systems give practical advice on future studies on adaptive network expansion and dynamic regularization scheduling.

\section{Discussion}

\subsection{Theoretical Insights and Methodological Contributions}

The framework sets three important theoretical contributions in the study of continual learning. Neural Tangent Kernel condition numbers have been shown to be useful as predictive measures of capacity exhaustion to offer principled early warning indicators instead of heuristic thresholds. The critical numbers begin to occur at condition numbers that are greater than ten to the eleventh power, and catastrophic failure occurs at condition numbers that are greater than ten to the twenty-fourth power. Wilson confidence intervals allow strict statistical verification of the discovered computation paths such that only paths whose performance has statistical significance are granted protection. The multi-metric path quality measure that incorporates performance, stability, gradient importance, magnitude of activation, and recency offers strong selection mechanisms that are better than single-metric based methods.

A paradoxical effect in the dynamics of forgetting appears: against what one might have assumed of increasing forgetting, the trend in the observed pattern is one decaying forgetting throughout task sequences. This implies that continual learning systems can be self-stabilized via cumulative protection due to validated paths, ideal replay buffer size and effective allocation of plasticity. These results are contradictory to the general wisdom about the accumulation of catastrophic forgetting and open research perspectives in learning about positive transfer and emergent equilibrium behavior in sequential learning systems.

\subsection{Capacity Limits and Future Research Directions}

The task sequence limits analysis helps provide insights into the ultimate barriers of fixed-capacity continual learning architectures. The three mechanisms that occur independently to show simultaneous capacity exhaustion are frozen parameter accumulation that reaches eighty percent of network capacity, Neural Tangent Kernel collapse that approaches numerical instability and regularization domination that directs eighty-one percent of learning capacity towards prior tasks. These convergent pointers set viable task sequence limits in fixed-capacity networks with outright implications to deployment conditions.

Future research directions are expansion of the network dynamically by condition number thresholds, instead of by manual tuning and adaptive regularization scheduling which decreases as plasticity is utilized, as well as selective replay with importance sampling which prioritizes task-representative examples. There are three practical limitations which ought to be mentioned. To start with, the size of memory is proportional to the amount of tasks. Second, the computation of Fisher information matrices is computationally expensive. Third, the process needs clear task identifiers for applicable in fully continual environments.These limitations suggest the possibility of memory-efficient architectures, amortized computation schemes, and unsupervised task discovery schemes of lifelong learning systems.

	\section{Conclusion}
	
A path-coordinated continual learning system is introduced by combining the NTK-justified plasticity adaptation, statistical path validation with the Wilson confidence interval, and evaluation of path quality as measured in multi-metrics. This approach obtains an averages 66.7 percent accuracy with 23.4 percent forgetting on Split-CIFAR10, which means that it is competitive in performance and preserves theoretical rigor and complete reproducibility. The analysis has yielded three important conclusions: (1) NTK condition numbers predict that capacity exhaustion occurs at $>10^{11}$ and provide principled early warning signs; (2) the reduction of forgetting across task sequence (27\% → 17\%) is an indicator of novel self-stabilization processes in continual learning systems; (3) the establishment of critical thresholds (replay loss 1.5-3.5, freeze factor $<$0.35, condition number $<10^{11}$), allows practitioners to be guided. A comprehensive examination of the Task 4 failure presented in the paper gives practical insights into the future studies of adaptive capacity management, dynamical network expansion, and hierarchical regularization of continual learning systems. The model builds mathematical principles on continual learning design that goes beyond the particular architectural decisions in this work.

\end{document}